\documentclass[journal]{IEEEtai}

\usepackage{booktabs}
\usepackage{adjustbox}
\usepackage{booktabs}
\usepackage{tabularx}
\usepackage{threeparttable}
\usepackage[colorlinks,urlcolor=blue,linkcolor=blue,citecolor=blue]{hyperref}
\usepackage{amsmath}
\usepackage{amsthm}
\usepackage{amsfonts}
\usepackage{amssymb}
\usepackage{bm}
\usepackage{epsfig}
\usepackage{blindtext,graphicx}
\usepackage{multicol}
\usepackage{epstopdf}
\usepackage{subcaption}
\usepackage{booktabs}
\usepackage{array}
\usepackage{multirow}
\usepackage[backend=bibtex,bibstyle=ieee,citestyle=numeric-comp]{biblatex}
\usepackage{xcolor,soul,framed}
\usepackage{balance}
\DeclareMathAlphabet\mathbfcal{OMS}{cmsy}{b}{n}
\bibliography{refs}
\usepackage{color,array}

\usepackage{graphicx}
\usepackage{booktabs}

\begin{document}

\title{Histogram Layers for Neural ``Engineered" Features}


\author{Joshua~Peeples,~\IEEEmembership{Member,~IEEE}, Salim Al Kharsa,  Luke~Saleh,~\IEEEmembership{Student Member,~IEEE}
      and Alina~Zare,~\IEEEmembership{Senior Member,~IEEE}
\thanks{Manuscript received July 22, 2024.; revised February 27, 2025; accepted July 23, 2025.}
\thanks{J. Peeples is an Assistant Professor and S. Al Kharsa is an undergraduate student in the Department of Electrical and Computer Engineering, Texas A\&M University, College Station, TX, 77845, USA, e-mails: jpeeples@tamu.edu and k1411130@tamu.edu.}
\thanks{L. Saleh is an undergraduate student and A. Zare is a Professor in the Department of Electrical and Computer Engineering, University of Florida, Gainesville, FL, 32608, USA, emails: lukesaleh@ufl.edu and azare@ece.ufl.edu.}} %

\markboth{Accepted to IEEE Transactions on Artificial Intelligence}
{Peeples, \MakeLowercase{\textit{et al.}}: Histogram Layers for Neural ``Engineered" Features}


\maketitle

\begin{abstract}
 In the computer vision literature, many effective histogram-based features have been developed.   These ``engineered" features include local binary patterns and edge histogram descriptors among others and they have been shown to be informative features for a variety of computer vision tasks. In this paper, we explore whether these features can be learned through histogram layers embedded in a neural network and, therefore, be leveraged within deep learning frameworks. By using histogram features, local statistics of the feature maps from the convolution neural networks can be used to better represent the data.  We present neural versions of local binary pattern and edge histogram descriptors that jointly improve feature representation for image classification. Experiments are presented on benchmark and real-world datasets. Our code is publicly available \footnote{\url{https://github.com/Advanced-Vision-and-Learning-Lab/NEHD_NLBP}}.

\end{abstract}

\begin{IEEEImpStatement}
 Our research introduces learnable histogram-based features in neural networks, such as neural local binary patterns (NLBP) and neural edge histogram descriptors (NEHD). These novel layers enable the integration of statistical image processing algorithms directly into deep learning frameworks. By leveraging these learnable histogram-based features within deeper networks, we can enhance feature representation capabilities, thereby improving performance in image classification tasks and potentially discovering new histogram-based features.  Our publicly available code facilitates broader adoption and exploration of these techniques, contributing to advancements in computer vision applications that rely on robust, adaptive, and explainable feature extraction methods.

\end{IEEEImpStatement}


\section{Introduction}
\label{sec:intro}
Before the popularity of deep learning, feature engineering often (and still does) played a vital role in the fields of computer vision and machine learning. Examples of engineered features include local binary patterns \cite{ojala1994performance} (LBP) and edge histogram descriptors (EHD) \cite{manjunath2002introduction}. Extracting and selecting optimal features through feature engineering was often difficult. Many of these engineered features were created through the use of histograms, but there were many parameters (such as bin centers and widths) that were difficult to tune. As an alternative to the difficult and time-consuming process of feature engineering, deep learning is used to automate the process of extracting features and performing follow-on tasks (\textit{e.g.}, classification, segmentation, object detection). 

Convolutional neural networks (CNN) are frequently used in a variety of applications. These models extract features through a series of convolutions, aggregation functions (\textit{e.g.}, max and average pooling) and non-linear activation functions to improve the representation of the data for better performance. Despite these powerful expressive features learned by a CNN, these models have some downfalls. CNNs have a spatial stationary assumption; therefore, CNNs are unable to account for changes in the local statistics (\textit{i.e.}, statistical texture features \cite{peeples2022histogram}) for regions of the data \cite{taigman2014deepface,liang2021patch}. Additionally, CNNs can lead to increased computational cost in terms of training time, memory requirements, and large datasets \cite{juefei2017local}. CNNs effectively capture structural texture features while histogram layers focus on statistical texture features \cite{peeples2022histogram}.

To mitigate these issues associated with traditional and deep learning features, alternative models have been introduced that take inspiration from both approaches. A notable method is the local binary convolutional neural network (LBCNN) \cite{juefei2017local}. The LBCNN used a novel architecture design that led to a less expensive model that performed comparably to standard CNNs. LBCNN was inspired by LBP; however, the LBCNN generalizes the code generation and does not account for the aggregation operation of the original LBP.

 \begin{figure}[t]
    \centering
    \includegraphics[width=1\linewidth]{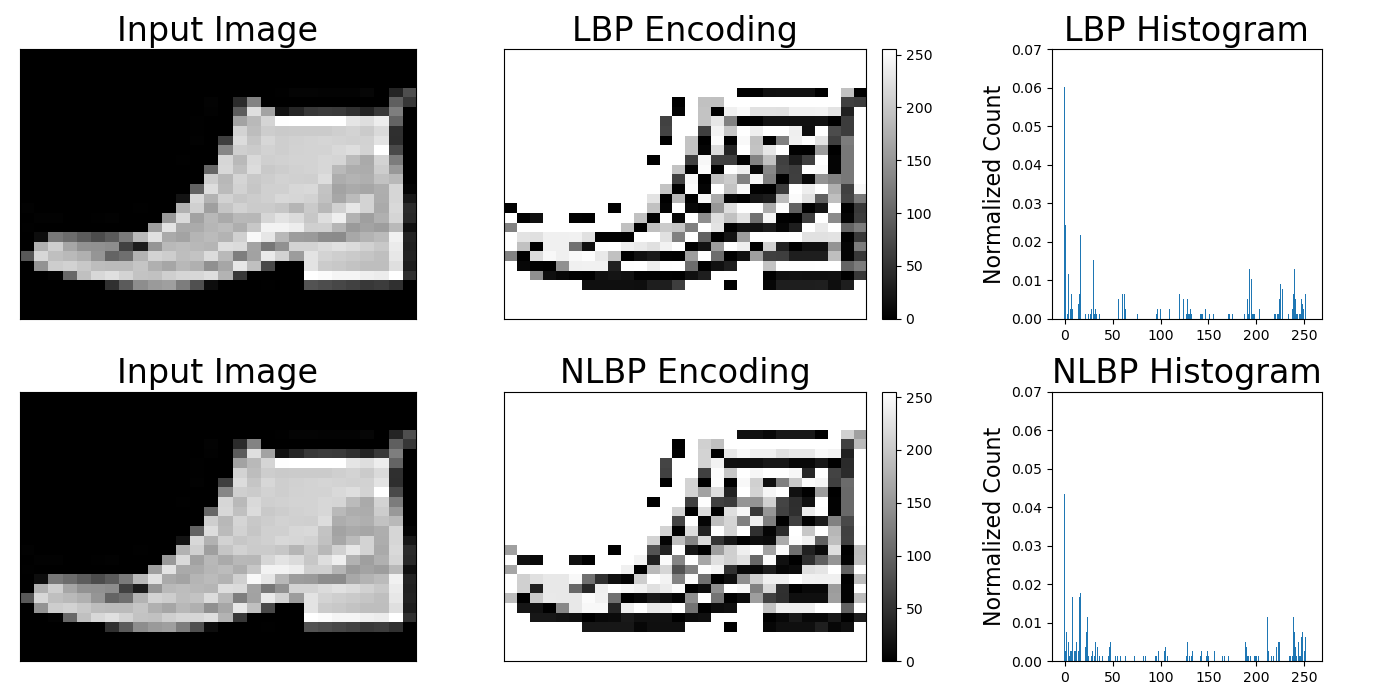}
    \caption{Feature maps captured by LBP (top) and NLBP (bottom) from the FashionMNIST dataset. The proposed approach nearly reconstructs the original LBP feature. The LBP and NLBP encodings differ slightly due to the non-linearity: threshold (LBP) and ReLU (NLBP). Additionally, a square window ($3 \times 3$) is used for NLBP while LBP used a radial window for the local neighborhood. There are small magnitude differences of the histogram because of the soft binning approximation used in NLBP.}
    \label{fig:LBP_Final_Reconstruction}
\end{figure}

 \begin{figure*}[htb]
    \centering
    \includegraphics[width=1\linewidth]{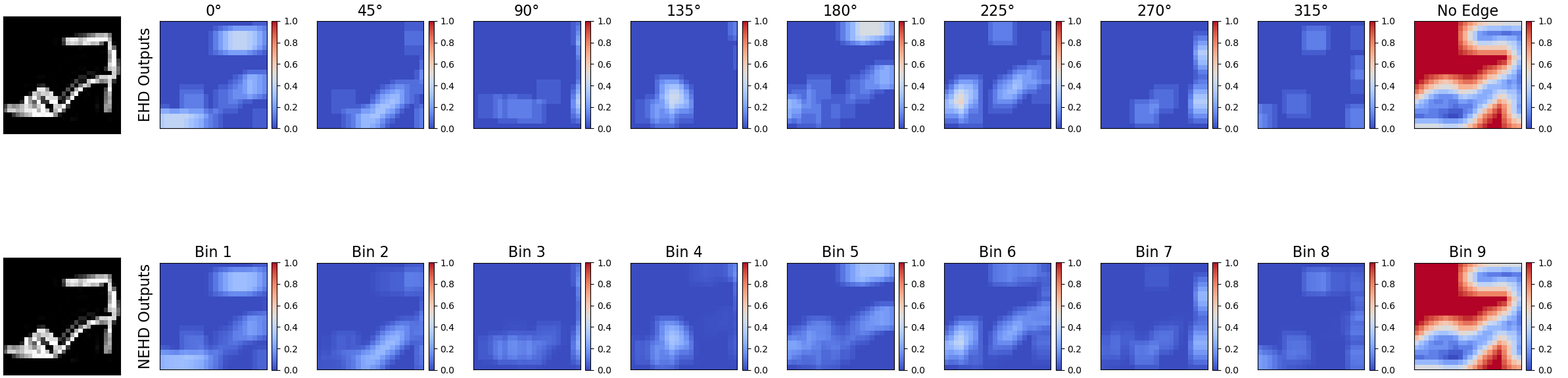}
    \caption[EHD Final Reconstruction]{Features maps captured by EHD (top row) and NEHD (bottom row) from the FashionMNIST dataset. The proposed approach closely reconstructs the original EHD feature. There are small magnitude differences because of the soft binning approximation used in NEHD.}
    \label{fig:EHD_Final_Reconstruction}
\end{figure*}

Histogram-based feature approaches (\textit{e.g.}, LBP and EHD) as well as CNNs are able to extract texture information \cite{maenpaa2005texture,liu2019bow,hermann2020origins,peeples2021histogram} to represent the data. There are two types of textures: statistical and structural \cite{peeples2021histogram,ji2022structural}. Both texture feature types are useful to represent the data to improve performance and features such as LBP use both within the feature extraction pipeline \cite{maenpaa2005texture}. In this work, we present a generalized framework to learn histogram-based features in artificial neural networks: neural LBP (NLBP) and EHD (NEHD). The contributions of this new method are four-fold:
\begin{itemize}
    \item Flexibility: mitigation of issues associated with parameter selection
    \item Expressibility: ability to change feature representation to best match problem (original LBP/EHD may not be best for problem)
    \item Synergy: fusion of statistical and structural textures in network design for neural ``engineered" features
    \item Utility: the proposed layers can be used to learn histogram-based features and in future work, other powerful histogram-based features could be discovered through training this layer.
\end{itemize}

Each neural feature can be reconstructed using convolutional and histogram layers to learn the structural and statistical texture features respectively. Our proposed neural ``engineered" features can closely reconstruct the LBP and EHD features as shown in Figures \ref{fig:LBP_Final_Reconstruction} and \ref{fig:EHD_Final_Reconstruction}. The LBP feature consists of both structural (\textit{i.e.}, encoded pixel differences) and statistical (\textit{i.e.}, binary code frequency) texture information. To capture the structural relationships, following \cite{liu2017local}, sparse convolutional kernels can be used to capture the difference between the center and neighboring pixels in an image. After the convolution operation is performed, an activation function (\textit{e.g.}, sigmoid, ReLU) can be used instead of the threshold operation. After the input image has been encoded, a histogram layer can be used to adaptively aggregate the encoded features as opposed to the fixed binning of the original LBP feature. 

Figure \ref{fig:LBP_Final_Reconstruction} shows a comparison of LBP and the proposed NLBP. To reconstruct the LBP feature, a modified ReLU activation function can be used. The standard ReLU function sets all negative values to $0$ (similar to the threshold function in LBP), but if the center pixel and neighbor pixel value are the same, the LBP threshold function will output a value of $1$. To account for this in the NLBP, if the difference between a center and neighboring pixel is $0$, the ReLU function was set to output a $1$ instead of a $0$ resulting in a similar encoding map as the original LBP. The histogram layer bins are initialized to the corresponding LBP code values ranging from $0$ to $255$. The bin width was set to $3.75$ as this value will correspond to a narrow histogram bin.

The EHD feature also captures structural (\textit{i.e.}, edges) and statistical (\textit{i.e.}, max response frequency) texture information. The structural features can be captured using a convolutional layer with each filter corresponding to a Sobel kernel for each edge orientation, $0^{\circ}$ to $315^{\circ}$. A total of eight edge kernels were used to capture horizontal, vertical, diagonal and anti-diagonal orientations (as in the standard EHD feature). To account for direction information, the bin center values are set to the maximum response of the convolution output from the edge kernels and the input data (\textit{i.e.}, maximum input value $\times$ ($1 + 2 + 1$)). By changing the bin center, the binning function will now account for direction information. The anti-direction will now have the lowest response to the corresponding correct direction (\textit{i.e.}, $0^\circ$ will have lowest response for $180^\circ$). The same bin width value ($3.75$) was used for NEHD as in NLBP.
The parameters (convolution kernels, bin centers, and bin widths) were fixed to ensure that the histogram layer can mimic the EHD feature as shown in Figure \ref{fig:EHD_Final_Reconstruction}.

 The spatial location of the responses align properly as shown in Figure \ref{fig:EHD_Final_Reconstruction}. The no edge response of the NEHD and EHD feature maps are nearly the same. Since NEHD is a soft approximation of EHD, there will be some overlap between each edge orientation. As a result, there are small magnitude differences between each edge orientation. The bin width can be tuned to produce a more exact representation for each edge response. The EHD feature can be reconstructed with the histogram layer if 1) the weights on the input features represent edge kernels and 2) the bin centers are equal to the maximum response of each orientation.

\section{Related Work}

\begin{figure*}[htb]
    \centering
	\includegraphics[width=.99\linewidth]{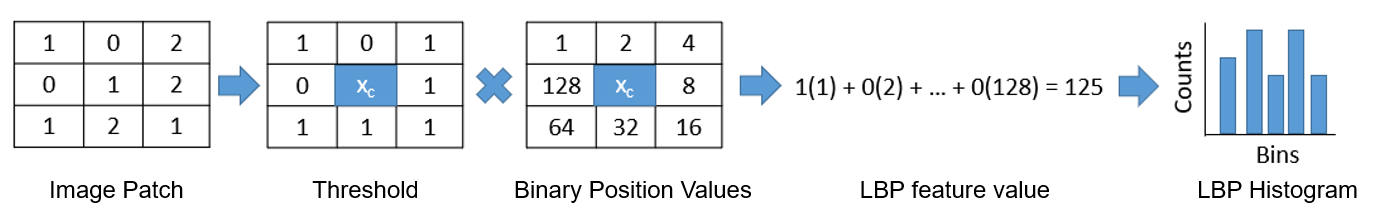}
	\caption{Example of LBP feature value calculation. After a local neighborhood is selected (\textit{e.g.}, $3 \times 3$), the difference between the center and neighboring pixels is computed. If the difference is greater than or equal to a chosen threshold (\textit{e.g.}, $0$), then a value of ``1" is returned for that neighbor. The next step is to encode the binary position of each neighbor (\textit{i.e.} $2^0$, $2^1$, ... , $2^7$). An elementwise multiplication and sum operation is performed between the threshold and binary position values. The output of this operation is the LBP feature for the local patch. The values from each local patch of the input image is then aggregated via a histogram to generate the output LBP histogram vector.}
	\label{fig:LBP_ex}
\end{figure*} 

\subsection{``Engineered" Histogram Features}
Histograms are used throughout computer vision and machine learning as a method to aggregate intensity and/or feature values as well as relationships between neighboring inputs (\textit{e.g.}, edge orientation, pixel differences). Common histogram-based features include LBP \cite{ojala1994performance}, histogram of oriented gradients (HOG) \cite{dalal2005histograms}, EHD \cite{frigui2008detection}, and {Haralick texture features extracted from the} gray level co-occurrence matrix (GLCM) \cite{haralick1973textural}. For this work, we will focus on LBP and EHD.

\paragraph{LBP Review}
LBP is computed by first selecting a neighborhood of size $\mathcal{N}$ (commonly 3 $\times$ 3 or also using a radial window operation) and computing the difference between a center pixel, $x_c$, and each of the pixels in the defined neighborhood, $x_i$, in a gray scale image. The LBP feature value for a pixel in an image is computed by computing the product of the threshold, $\mathcal{T}$, of the difference between the center pixel and neighbor pixel, and two raised to the power of the index of the neighbor (Equation \ref{Eqn:LBP1}). The threshold for assigning a ``0" or a ``1" is usually 0 (\textit{i.e}., the difference needs to be positive) as shown in Equation \ref{Eqn:LBP2}, but other variants may use different thresholds \cite{zhou2009face,jiang2016adaptive}:

\begin{equation}
LBP(x_c) = \sum_{i=0}^{\mathcal{N}-1}\mathcal{T}(x_i-x_c)2^i
\label{Eqn:LBP1}
\end{equation}

\begin{equation}
\mathcal{T}(x_i-x_c) = \begin{cases}
1, (x_i-x_c)\geq0\\
0, otherwise
\end{cases}
\label{Eqn:LBP2}
\end{equation}

A toy example of the LBP feature for a local region of an image is shown in Figure \ref{fig:LBP_ex}. Once the LBP code value is computed for each pixel in an image, a histogram of the code values with maximum LBP code value of $G$, $\mathbfcal{H} \in \mathbb{R}^{G+1}$, are summed over the spatial dimensions, $M \times N$, of the image as shown in Equation \ref{Eqn:LBP3} is used as the descriptor (following notation from \cite{guo2010completed}):

\begin{equation}
\mathcal{H}_g = \sum_{i=0}^{M}\sum_{j=0}^{N}f(LBP(x_{ij}),g), g \in [0,G]
\label{Eqn:LBP3}
\end{equation}

\begin{equation}
f(x,y) = \begin{cases}
1, x = y\\
0, otherwise
\end{cases}
\label{Eqn:LBP4}
\end{equation}

\noindent LBP is robust to monotonic changes in grayscale due to the threshold function \mbox{\cite{ojala1994performance,ojala1996comparative,humeau2019texture}} (\textit{i.e.}, if the intensity values of an image changes, the binary threshold will always return either two values, ``0" or ``1", despite gray level changes). There are a plethora of extensions to LBP in the literature {for different applications including understanding cell phenotypes leveraging microscopy images \mbox{\cite{fekri2021cell}} and for biomedical applications such as pap smear classification \mbox{\cite{fekri2019pap}}.}
Several works are available that summarize the novelty of different LBP-based approaches \cite{liu2017local,fernandez2013texture}. {Work continues to be done to make improvements for the LBP feature in particular and our method aims to contribute to ongoing work related to this feature}. 

\paragraph{EHD Review}
The EHD feature records the frequency and orientation of intensity changes in an image \cite{frigui2008detection}. A single channel input image is convolved with a set of $K$ filters (\textit{e.g.}, Sobel \cite{sobel19683x3}) of size $M \times N$ to calculate the edge responses, $R \in \mathbb{R}^{M' \times N' \times K}$. The edge responses are grouped (``binned'') together based on their orientation such as vertical, horizontal, diagonal, anti-diagonal, and no edges (isotropic) \cite{frigui2008detection}. The five groups can include signed \cite{peeples2018possibilistic,peeples2019comparison} or unsigned orientations \cite{frigui2008detection} (\textit{e.g.}, 0$^{\circ}$ and 180$^{\circ}$ are both horizontal but different directions). To compute the histogram of edge orientations, $\mathbf{H} \in \mathbb{R}^{K+1}$, the maximum edge responses are summed over the spatial dimensions for the $k^{th}$ element (\textit{i.e.}, bin) as shown in Equation \ref{Eqn:EHD1}: 

\begin{equation}
H_k = \sum_{i=0}^{M^\prime}\sum_{j=0}^{N^\prime}\mathcal{B}_k(R_{ijk})
\label{Eqn:EHD1}
\end{equation}

In order to only keep ``strong" edge responses, a global threshold, $\theta_G$, is selected and the voting function, $\mathcal{B}$, transforms the edge responses to ``votes" for each orientation as shown in Equations \ref{Eqn:EHD2} and \ref{Eqn:EHD3}. The first case to consider is when an edge response is to be recorded. The edge response must be larger than any other edge orientation and greater than or equal to the threshold. If both conditions are met, then a ``1" is assigned to the corresponding edge bin as shown in Equation \ref{Eqn:EHD2}:

\begin{equation}
\mathcal{B}(R_{ijk})= \begin{cases}
0, \underset{l: k \neq l}{\exists} R_{ijl} > R_{ijk} \text{ and } R_{ijk}\geq\theta_G\\
0, R_{ijk} < \theta_G\\
1, otherwise
\end{cases}
\label{Eqn:EHD2}
\end{equation}

\begin{figure*}[htb]
    \centering
	\includegraphics[width=.9\linewidth]{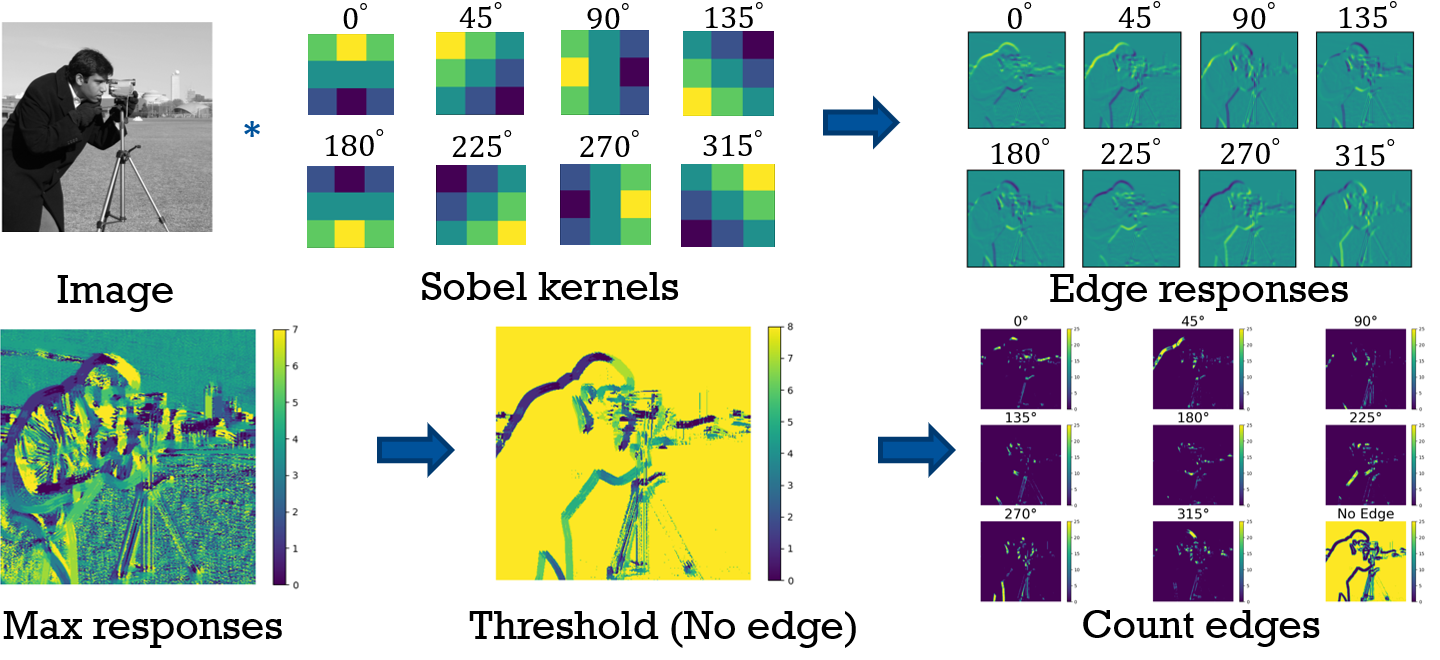}
	\caption{Overall process for EHD feature. The image is first convolved with Sobel kernels \cite{sobel19683x3} to generate edge responses for each orientation. The next step is to record the orientation that has the max response. Once the max response is recorded, a threshold is applied to ensure only ``strong" edges are retained. The final step is to aggregate the max responses for each edge orientation to produce the final feature maps. The edge counts are binned into a histogram feature vector (similar to LBP).}
	\label{fig:EHD_ex}
\end{figure*} 

\noindent The ``no edge" case is when all edge responses are lower than the global threshold. When this condition is met, a value of ``1" is assigned to the ``no edge" or isotropic orientation 
\begin{equation}
\mathcal{B}(R_{ijk+1})= \begin{cases}
1, \forall k = 1, ..., K; R_{ijk} < \theta_G\\
0, otherwise
\end{cases}
\label{Eqn:EHD3}
\end{equation}

An example of EHD features is shown in Figure \ref{fig:EHD_ex}. ``Engineered" histogram-based features such as LBP and EHD have the same limitations as traditional feature engineering approaches. These limitations include manually tuning parameters and domain expertise \cite{nanni2017handcrafted}. To overcome these issues, others have proposed to 1) combine traditional and deep learning feature extraction approaches and 2) design neural networks to extract ``engineered" features through network design. 

\begin{figure*}[htb]
	\begin{subfigure}{1\textwidth}{
			\includegraphics[draft=false,width=\textwidth]{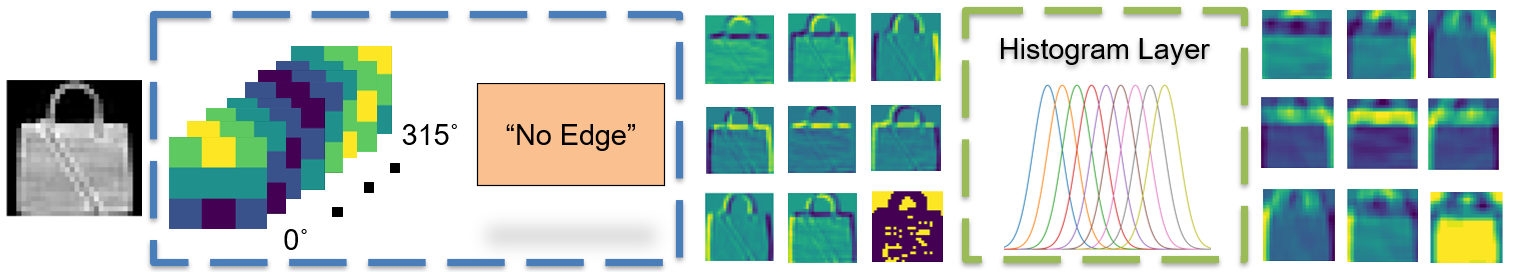}
			\caption{NEHD Implementation}
			\label{fig:NEHD_model}
		}
	\end{subfigure} 
 
	\begin{subfigure}{1\textwidth}{
			\includegraphics[draft=false,width=\textwidth]{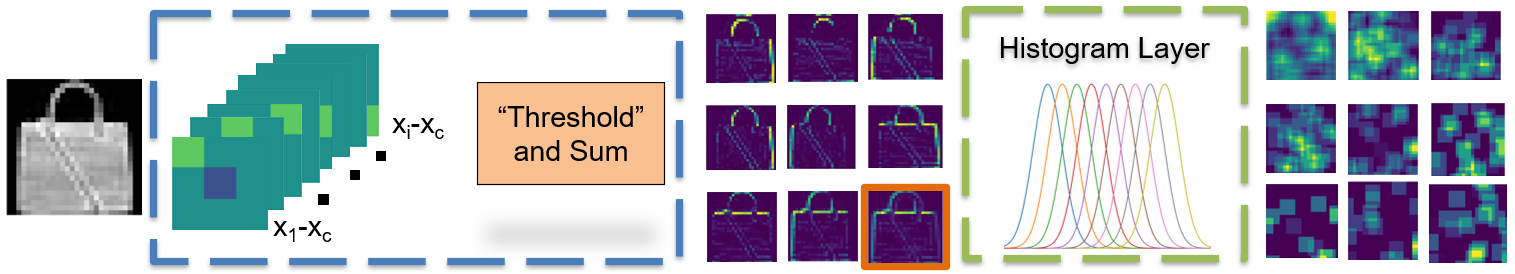}
			\caption{NLBP Implementation}
			\label{fig:NLBP_model}
		}
	\end{subfigure}
	\caption{The proposed neural ``engineered" feature pipeline for NEHD (Figure \ref{fig:NEHD_model}) and NLBP (Figure \ref{fig:NLBP_model}). The structural texture information is captured in the blue box and the statistical texture information is shown in the green box. In Figure \ref{fig:NEHD_model}, the input image is convolved with kernels to capture each edge orientation. The ``no edge" operation can be implemented using a threshold or convolution layer. In Figure \ref{fig:NLBP_model}, the input image is convolved with kernels to capture the difference between neighboring pixels and the center pixel. The ``threshold" operation is implemented using an activation function (\textit{e.g.}, sigmoid). After the ``threshold" function, a $1 \times 1$ convolution is used to compute a weighted sum operation of the difference maps to generate the ``bit" map of LBP highlighted in orange. After the structural texture information is extracted, both features use a histogram layer to aggregate the structural texture information into statistical texture features.}
	\label{fig:model}
\end{figure*} 

\subsection{Combination of Traditional and Deep Learning Approaches}
Traditional feature extraction approaches and deep learning methods are used 1) separately and 2) together. Generally, there are five approaches for combining traditional and deep learning approaches \cite{peeples2022connecting}: 1) take features from deep learning model and pass into traditional classifiers \cite{scabini2019evaluating,sani2017learning,zhang2016svm}, 2) ``engineer" filters of CNN (may update through backpropagation or keep fixed) or emulate ``engineered" features via the network design \cite{bruna2013invariant,chan2015pcanet,malof2018improving,bianconi2019cnn,Su_2021_ICCV} , 3) pass ``engineered" features into the network as input \cite{muhammad2017tex,anwer2018binary,van2019feeding}, 4) combine both ``engineered" and CNN features for traditional classifiers \cite{nguyen2018combining,paul2016combining,wu2016multi}, and 5) texture encoding methods \cite{cimpoi2015deep,zhang2017deep,song2017locally,xue2018deep,hu2019multi}. 

There are several problems with existing approaches for combining traditional and deep learning methods. First, an issue with incorporating ``engineered" features are that the models cannot be trained in an end-to-end fashion because the ``engineered" features are not updated as performance changes. Also, there are additional computational costs for training deep learning model(s) and separate classifier(s). Along with the increased computational costs, more parameter tuning is necessary for the ``engineered" features, deep learning model(s), and classifier(s).  Lastly, deep learning features perform well in practice, but these features are not easily explainable and/or interpretable as traditional features though there are ongoing efforts to ``open" the black box \cite{rudin2022interpretable}. 


\label{sec:related_works}

\section{Method}
\label{sec:method}
\subsection{Neural ``Engineered" Features Histogram Layer}
    
    The baseline histogram layer function in \cite{peeples2021histogram} equally weights input features. As constructed, this implementation of the histogram layer will not be able to account for structural changes in texture information. For example, two binary images of a cross and checkerboard pattern will have the same distribution of pixels, but the arrangement of these pixels is different \cite{peeples2022histogram}. The unweighted histogram layer will misidentify these two different structural texture types. To account for this, one can simply learn a weighting of the input features to account for structural differences. Our proposed neural ``engineered" feature layer, $f$, takes an input image or feature map(s), $\mathbf{X}$, and applies two functions to extract texture information as shown in Equation \ref{eqn:feat_extract}:
    
    \begin{equation}
        f(\mathbf{X}) = \phi\left(\sum_{\rho \in \mathcal{N}}^{} \psi(\mathbf{x}_\rho)\right) \label{eqn:feat_extract}
    \end{equation}
    
    \noindent where $\phi$ and $\psi$ represent the statistical and structural texture information, respectively. Generally, $\psi$ is implemented as a local feature extractor in a given neighborhood $\mathcal{N}$ (\textit{e.g.}, $3 \times 3$ edge kernel) and $\phi$ is selected as a global or local operation.
    
    $\phi$ represents the histogram layer introduced in \cite{peeples2021histogram}. The histogram layer output is shown in Equation \ref{equation:Weighted_RBF}. The normalized frequency count, $\phi_{rcbk}$, is computed with a sliding window of size $S \times T$ and the binning operation for a histogram value in the $k^{th}$ channel of the input $x$:
    \begin{equation}
    \phi_{rcbk} =  
    \cfrac{1}{ST}\sum_{s=1}^{S}\sum_{t=1}^{T}\exp\left(-\gamma_{bk}^2\left(\sum_{\rho \in \mathcal{N}}^{} \psi_{bk}(\mathbf{x}_\rho)-\mu_{bk}\right)^2\right)
    \label{equation:Weighted_RBF}
    \end{equation}
    where $r$ and $c$ are spatial dimensions of the histogram feature maps. The histogram layer is used to aggregate the structural texture information from the input. We detail the structural texture feature extraction process for NEHD and NLBP in the next two sections.  
    
\subsection{Neural Edge Histogram Descriptor} 

NEHD captures structural texture information by accounting for edge information. $\psi^{NEHD}_{bk}$ is shown in Equation \ref{eqn:NEHD} where the input, $x$, is weighted by $w_{mnk}$ corresponding to a value in an $M \times N$ kernel for each input channel is defined as:
    \begin{equation}
       \psi^{NEHD}_{bk} = \sum_{m=1}^{M}\sum_{n=1}^{N}w_{mnk}x_{r+s+m,c+t+n,k}
        \label{eqn:NEHD}
    \end{equation} 
Equation \ref{eqn:NEHD} corresponds to the edge responses, $\mathcal{R}$, discussed in Equation \ref{Eqn:EHD1}. Once the structural texture is captured, the output from $\psi^{NEHD}$, is then passed into the histogram layer. Unlike the baseline EHD feature, the NEHD can be trained end-to-end to update both the structural and statistical texture representation. 

The NEHD edge responses can be easily implemented using a convolutional layer to capture the edge feature maps as shown in Figure \ref{fig:NEHD_model}. The histogram layer is also implemented using two convolutional layers for the binning of the edge responses. For the ``no-edge" orientation, two approaches can be used: thresholding and convolution. Similar to EHD, after the edge responses are computed from Equation \ref{eqn:NEHD}, a threshold can be applied to detect if there is a ``strong" edge present. The ``no-edge" map could then be simply concatenated to the edge feature maps from the convolutional layer. The second approach is to learn the thresold operation by using a $1 \times 1$ convolution to map the edge response maps to a single channel. A non-linearity (\textit{e.g.}, sigmoid) is then applied to the single feature map to learn a differentiable threshold operation. We investigate both approaches in Section \ref{sec:results_discussion}.

\subsection{Neural Local Binary Pattern} 

NLBP captures structural texture information by accounting for pixel differences. Similar to the LBCNN approach \cite{liu2017local},  $\psi^{NLBP}_{bk}$ is shown in Equation \ref{eqn:NLBP} where the input, $x$, is weighted by $w_{mnk}$ corresponding to a value in an $M \times N$ kernel for each input channel is

\begin{equation}
   \psi^{NLBP}_{bk} = \sum_{z=1}^{Z}\sigma\left(\sum_{m=1}^{M}\sum_{n=1}^{N}w_{mnk}x_{r+s+m,c+t+n,k}\right) \mathcal{V}_{kz}
    \label{eqn:NLBP}
\end{equation} 
where $\sigma$ is an activation function (\textit{e.g.}, sigmoid, ReLU) and $\mathcal{V}_{kz}$ is a learnable $1 \times 1$ convolution. Equation \ref{eqn:NLBP} parallels Equations \ref{Eqn:LBP1} and \ref{Eqn:LBP2}. The difference between a neighbor pixel and center pixel can be implemented using sparse convolution kernels where the weight on the center pixel is $-1$, a neighbor pixel is 1, and all other values in the kernel are $0$ as shown in Figure \ref{fig:NLBP_model}. This operation can be extended to other neighbors by rotating the kernel to process all pixels in the defined neighborhood. After the difference is computed, instead of the threshold operation, an activation function such as sigmoid can be used to map the values between $0$ and $1$. The final step is to multiply the ``thresholded" difference maps by the binary base. This can also be implemented using the $1 \times 1$ convolution where the base value can be learned and not fixed to be a power of $2$. 

\subsection{Network Architecture}

\label{sect:network_architecture}
{A layer is a sequence of operations in an artificial neural network that process an input and produce an output. For example, a convolution layer consists of convolution (affine transformation), non-linearity (detector stage), and pooling layer as discussed in \mbox{\cite{goodfellow2016deep}}. The ``layers” in our approach are the structural (convolution and non-linearity) and statistical (histogram layer \mbox{\cite{peeples2021histogram}} which is comprised of a convolution, non-linearity, and pooling layers) texture feature extractors. For this work, a simple network was used to evaluate the quality of the features learned. The input image is passed into either the NEHD or NLBP layer as shown in Figure \mbox{\ref{fig:model}}. The models in this work are comprised of the input layer (i.e., input images), two feature extraction layers (structural and statistical textures), and an output layer for classification. For the structural texture layers for NEHD and NLBP, the size of the kernel and/or dialation can be used to adjust the receptive field. The statistical texture layers leverage local histogram layers and the aggregation neighborhood can also be adapted similarly to the structural layer.} 

\subsection{Computational Complexity Analysis}
{Each method has a similar computational complexity when extracting structural and statistical features. Given $C$ is the product of the number of input ($C_{in}$) and output ($C_{out}$) channels ($C = C_{in}C_{out})$, $K$ as the product of the kernel size ($K = MN$), and spatial input size of $H \times W$, the computational complexity of the structural features is equal to $\mathcal{O}(CKHW)$. This is due to the two convolutional layers used for both NEHD and NLBP. For NEHD, the first convolution layer is used to learn the edge orientations and a second convolution is used to learn the no edge orientation. While with NLBP, the first convolution is used to learn the neighborhood differences along with a second convolution to learn a weighted sum of the pixel differences. Both methods also use a histogram layer for the statistical features which uses two $1 \times 1$ convolution layers to learn the bin centers and widths respectively followed by an  average pooling layer resulting in a computational complexity of $\mathcal{O}(CHW)$.}

\section{Experimental Setup}
\label{sec:experimental_setup}
\begin{figure*}
    \centering
    \includegraphics[width=.70\linewidth]{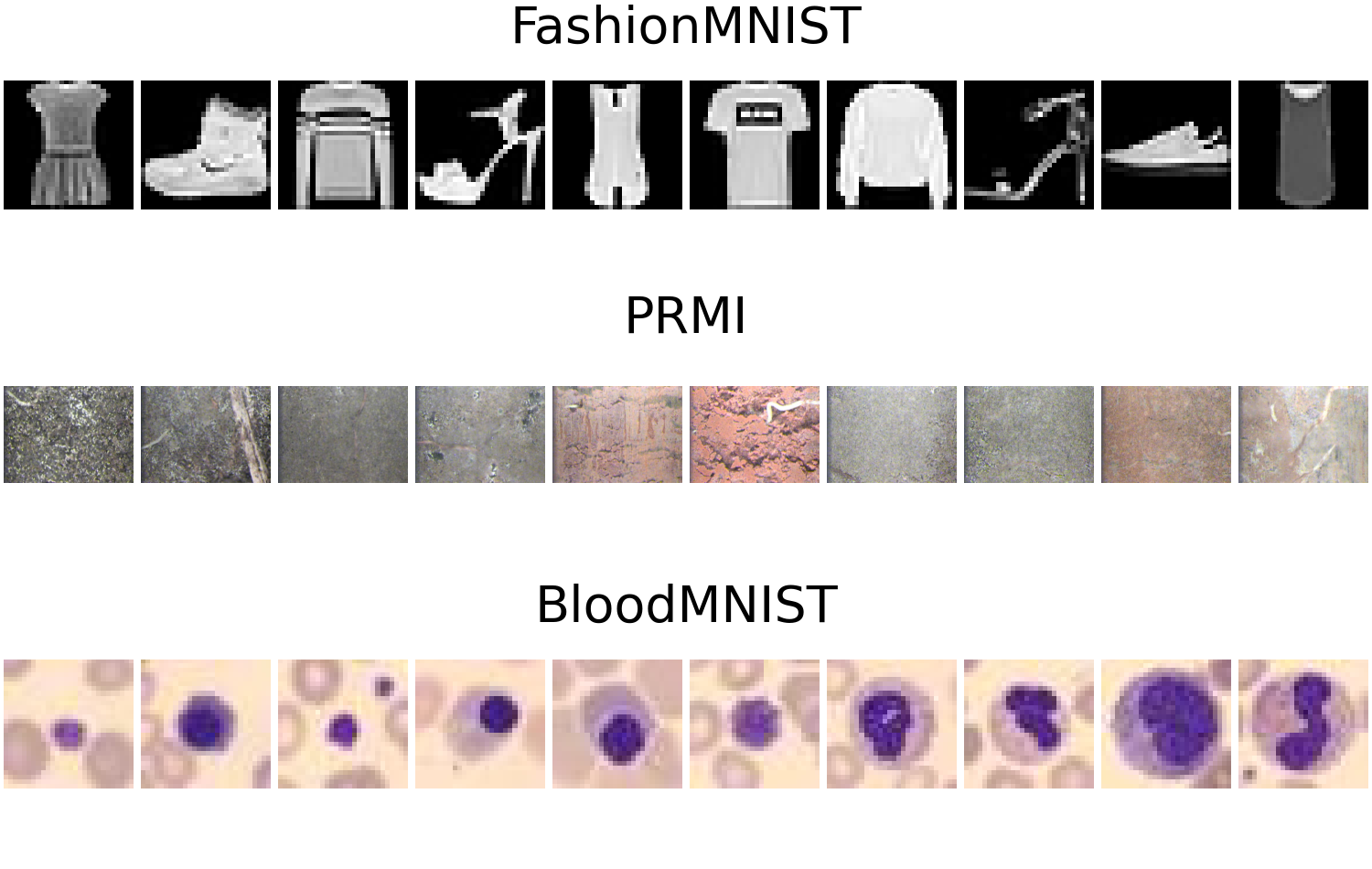}
    \caption{Example images from each dataset used in this work.}
    \label{fig:datasets}
\end{figure*}

The experiments are divided into two main parts: ablation study and feature comparison. All experiments used a shallow model {as described in Section \mbox{\ref{sect:network_architecture}}} that consisted of the input image, feature extraction (EHD, LBP, NEHD, and NLBP), and a fully connected layer. We chose this simple model to focus on the discriminative power of each feature using a linear classifier. In the ablation study, the FashionMNIST \cite{xiao2017fashion} dataset is used to investigate the impact of a) initialization and b) parameter learning in Section \ref{sec:ablation_init}. For the initialization experiments, the impact of initializing the layer randomly or with the baseline feature setting (\textit{i.e.}, NEHD kernels initialized with Sobel kernels and threshold operation) was evaluated. The feature learning focused on the contribution of learning a) statistical texture features, b) structural texture features, and c) both. 

{For NLBP and NEHD, we also investigated the impact of neighborhood size for the structural and statistical features in Seciton \mbox{\ref{sec:neighborhood}}. The kernel sizes for both structural features for NEHD and NLBP can be increased, but to stay as close to the original features as possible as well as reduce computational costs by not introducing more learnable parameters with larger kernels, the structural kernels were dilated for three sizes: 1, 2, and 3. This allowed us to evaluate the impact of a larger neighborhood size while retaining the properties of the original EHD/LBP features. For the statistical features, the aggregation kernel was varied from $3 \times 3$, $5 \times 5$, and $7 \times 7$.}

Additionally, texture features such LBP and EHD have been typically applied to grayscale or single channel inputs \cite{porebski2008haralick}. The Plant Root Minirhizotron Imagery (PRMI) dataset \cite{xu2022prmi} consists of RGB plant images. For both NEHD and NLBP, different multichannel approaches were investigated in Section \ref{sec:ablation_channels}: treat each channel independently, use $1 \times 1$ convolution to map an RGB image to single channel input, and converting an RGB image to grayscale. 

An extensive comparison is performed on three datasets (FashionMNIST \cite{xiao2017fashion}, PRMI \cite{xu2022prmi}, BloodMNIST \cite{medmnistv1,medmnistv2,bloodmnist}) with the baseline EHD and LBP, LBP variants, NEHD, NLBP, and {two lightweight CNNs, dilated spatial attention network (DSANet)\mbox{\cite{bhowmik2024dsanet}} and multi-scale dilated CNN (MSDCNN) \mbox{\cite{bhowmik2024utilization}}}  as discussed in Section \ref{sec:comparison}. {Example images from each dataset are shown in Figure \mbox{\ref{fig:datasets}}.} EHD was implemented using Pytorch. In order to compare our proposed method with the original LBP and LBP variants, the delayed function from the Dask package \cite{rocklin2015dask} was used to parallelize the Sklearn LBP feature extraction for each mini-batch of images. Each mini-batch is passed through a loop that holds each image to be computed in memory. Then, the Dask function produces a new list where the objects are all computed at the same time, and this list is converted to a tensor of the appropriate shape. This method allowed for faster processing speeds to compute the LBP of a mini-batch of images and integrate the features into the Pytorch framework. 

 BloodMNIST and PRMI have dedicated training, validation, and test partitions.  For FashionMNIST, the training data was divided into a 90/10 train and validation split. The trained model was then applied to the holdout test set. A subset of the PRMI (following \cite{peeples2022divergence}) was chosen using the following four classes: cotton, sunflower, papaya, and switch grass. No data augmentation was used for FashionMNIST. For BloodMNIST and PRMI, the data augmentation procedure followed \cite{peeples2021histogram,xue2018deep}. The images were resized to $128 \times 128$ with random and center crops of $112 \times 112$ as input into the models. The experimental parameters were the following: 100 epochs, batch size of 64, Adam optimization, and initial learning rate of $.01$ for EHD, NEHD, and NLBP. The initial learning rate for the LBP baseline and variants was $0.1$. {Following previous work \mbox{\cite{bhowmik2024dsanet,bhowmik2024utilization}}, DSANet and MSDCNN used an initial learning rate of $0.001$}. The EHD and NEHD parameters were set to the following: $3 \times 3$ edge kernel, $5 \times 5$ window size to aggregate bin counts, threshold of $0.9$, normalization of count (\textit{i.e.}, average pooling), and normalized kernel values. Both EHD and NEHD were set to extract eight edge orientations and ``no edge" resulting in a $9$-bin histogram. 
 
 The baseline LBP and variants used a radius of 1 and neighborhood size ($P$) set to $8$. Each LBP approach used the default number of bins: baseline LBP {\mbox{\cite{ojala1994performance}}} (256), uniform LBP {\mbox{\cite{ojala2002multiresolution}}} (59), {non-rotational invariant} (NRI)-uniform LBP {\mbox{\cite{ahonen2004face,ahonen2006face}}} (59), {rotation-invariant} (ROR) LBP {\mbox{\cite{ojala1996comparative}}} (256), and {variance} (Var) LBP {\mbox{\cite{ojala2002multiresolution}}} (256). NLBP used the following settings: $3 \times 3$ pixel difference kernel, $5 \times 5$ window size to aggregate bin counts, ReLU activation function, normalization of count (\textit{i.e.}, average pooling), and 16-bin histogram. The ReLU activation function was used in NLBP to promote better learning \cite{liu2017local} and the input range of the data was between $0$ and $1$ (range of the pixel difference is $-1$ and $1$ resulting in the ReLU function mapping the data between $0$ and $1$). A total of five experimental runs were completed for each model on the PRMI dataset while three experimental runs were used for BloodMNIST and FashionMNIST for each configuration. Experiments were completed on an A100 GPU.

\section{Results and Discussion}
\label{sec:results_discussion}

\subsection{Ablation Study: Initialization and Parameter Learning}
\label{sec:ablation_init}

\begin{table*}[htb]
    \centering
    \caption{FashionMNIST NEHD initialization and parameter learning mean test accuracy with $\pm$ 1 standard deviation across three experimental runs is shown. The baseline EHD approach had an average test accuracy of $86.94 \pm 0.00$. NEHD was initialized using a) EHD or b) randomly. The ``no-edge" was extracted by a)  learned using a convolution layer or b) threshold of edge response maps. The NEHD layer consisted of structural (convolution kernels) and statistical (histogram layer) texture features. The impact of updating both texture representations was captured across the columns of the table. The model with the best average test accuracy is bolded.}
        \begin{tabular}{|c|c|c|c|c|c|}
            \hline
            Random Initialization & No Edge Threshold & Learn Both & Learn Structural & Learn Statistical & Fix Both \\ \hline
                                    &                   & \textbf{89.74} \textbf{$\pm$} \textbf{0.12} & 89.09 $\pm$ 0.24 & 88.81 $\pm$ 0.05 & 83.82 $\pm$ 0.26 \\ \hline
                                    & \checkmark        & 89.51 $\pm$ 0.28 & 89.28 $\pm$ 0.24 & 88.35 $\pm$ 0.02 & 83.36 $\pm$ 0.06 \\ \hline
            \checkmark              &                   & 89.13 $\pm$ 0.23 & 88.87 $\pm$ 0.27 & 88.50 $\pm$ 0.17 & 78.08 $\pm$ 3.14 \\ \hline
            \checkmark              & \checkmark        & 89.16 $\pm$ 0.23 & 88.52 $\pm$ 0.22 & 88.28 $\pm$ 0.30 & 76.91 $\pm$ 2.54 \\ \hline
        \end{tabular}
    \label{tab:NEHD ablation}
\end{table*}

\begin{table*}
    \centering
    \caption{FashionMNIST NLBP initialization and parameter learning mean test accuracy with $\pm$ 1 standard deviation across three experimental runs is shown. The baseline LBP approach had an average test accuracy of $71.66 \pm 0.02$. NLBP was initialized using a) LBP or b) randomly. The NLBP base power was either a) fixed (power of $2$) or b) learned. The NLBP layer consisted of structural (convolution kernels) and statistical (histogram layer) texture features. The impact of updating both texture representations was captured across the columns of the table. The model with the best average test accuracy is bolded.}
    \begin{tabular}{|c|c|c|c|c|c|}
        \hline
        Random Initialization & Fixed Base & Learn Both & Learn Structural & Learn Statistical & Fix Both \\ \hline
                                &         & 85.54 $\pm$ 0.07 & 85.70 $\pm$ 0.01 & 86.63 $\pm$ 0.02 & 79.28 $\pm$ 0.00 \\ \hline
                                &   \checkmark                & 85.52 $\pm$ 0.04 & 85.56 $\pm$ 0.01 & 86.68 $\pm$ 0.02 & 78.65 $\pm$ 0.00 \\ \hline
        \checkmark              &         & 87.44 $\pm$ 0.40 & 86.67 $\pm$ 0.50 & 85.20 $\pm$ 0.03 & 81.84 $\pm$ 0.51 \\ \hline
        \checkmark              &   \checkmark                & \textbf{87.50} \textbf{$\pm$} \textbf{0.30} & 81.53 $\pm$ 0.42 & 80.74 $\pm$ 0.48 & 36.05 $\pm$ 1.60 \\ \hline
    \end{tabular}
    \label{tab:NLBP ablation}
\end{table*}

\subsubsection{NEHD}
The results of the NEHD initialization and parameter learning are shown in Table \ref{tab:NEHD ablation}. The proposed NEHD layer was robust to the random or EHD initialization as shown by overlapping error bars for each test accuracy except for when the structural and statistical textures were fixed. When the layer is fixed, the EHD initialization performance was more statistically significant than random initialization. Another interesting observation is that for most learnable texture features, the EHD initialization led to marginally better performance. Edges are important features that distinguish the classes in FashionMNIST as there is no background in the images and only the articles of clothing are present. Therefore, the edge ``profiles" for each clothing item can be a powerful discriminator between the different classes. This analysis is further validated as the baseline EHD feature achieved an average accuracy higher ($86.94$) than all of the fixed statistical and structural texture features models with random or EHD initialization. The EHD initialization for NEHD is not an exactly equal to EHD due to the soft binning approximation of the histogram layer.

The most difficult classes to distinguish were the ``T-shirt/top" and ``shirt." The baseline EHD had difficulty differentiating between these two classes as the representation is fixed and cannot adapt. However, our proposed NEHD layer outperformed the baseline EHD feature for learnable settings: learn structural, statistical, or both texture features. For the structural texture features, the edge kernel can be adapted from the initial Sobel kernels to account for more details regarding the structure of each clothing item. The statistical features assist with the soft binning approximation to mitigate small intra-class variations for each clothing item. As noted in Table \ref{tab:NEHD ablation}, the combination of learning both texture feature representations achieved the best performance on the FashionMNIST test set. 

\begin{table}[htb]
    \centering
    \caption{{Performance changes for NEHD and NLBP with varying dilation for structural features and aggregation window size for statistical features using the FashionMNIST dataset. The mean test accuracy with $\pm$ 1 standard deviation across three experimental runs is shown. The best average accuracy for each feature is bolded.}}
    \label{tab:hyperparams}
    \renewcommand{\arraystretch}{1.2} 
    \begin{tabular}{|c|c|c|c|}
        \hline
        \textbf{Dilation} & \textbf{Window Size} & \textbf{NEHD} & \textbf{NLBP} \\
        \hline
        1 & 3×3  & {89.49 $\pm$ 0.37} & 86.42 $\pm$ 0.39  \\ \hline
        1 & 5×5  & \textbf{89.74 $\pm$ 0.12} & \textbf{87.50 $\pm$ 0.30} \\ \hline
        1 & 7×7  & 87.82 $\pm$ 0.24 & 83.59 $\pm$ 0.44 \\
        \hline
        2 & 3×3  & 88.68 $\pm$ 0.24 & 86.82 $\pm$ 0.28 \\ \hline
        2 & 5×5  & 88.13 $\pm$ 0.12 & 85.41 $\pm$ 0.32 \\ \hline
        2 & 7×7  & 87.26 $\pm$ 0.23 & 84.30 $\pm$ 0.63 \\
        \hline
        3 & 3×3  & 88.05 $\pm$ 0.44 & {87.05 $\pm$ 0.17} \\ \hline
        3 & 5×5  & 87.00 $\pm$ 0.14 & 85.97 $\pm$ 0.21 \\ \hline
        3 & 7×7  & 85.63 $\pm$ 0.11  & 84.77 $\pm$ 0.45 \\
        \hline
    \end{tabular}
\end{table}

\subsubsection{NLBP} The results of the ablation study on NLBP are shown in Table \ref{tab:NLBP ablation}. The NLBP feature was robust to either random or LBP initialization except for the case when the kernel was randomly initialized and the base was fixed to be a power of $2$. This result matches similar analysis from \cite{liu2017local} that shows that fixing the base power may limit the generalization ability of the feature. Another interesting observation is that the randomly initialized models performed slightly better for every setting except when learning statistical features and fixing both texture feature parameters with a fixed base. For the statistical features, the initialization of the histogram layer to meaningful bin centers and widths will assist in the learning process of the model. For the structural textures, randomly initializing the kernels lead to an improvement in performance except for when the base power was fixed. This result indicates that the initialization of the kernel used to capture the relationship between neighboring pixels is important when capturing the structural texture information. The LBP initialized the NLBP feature with sparse kernels and this may cause limit the learning ability of the model. 

Similar to NEHD, maximal performance is achieved when both structural and statistical texture features are jointly learned. When LBP initialization is used, updating the statistical features achieved higher performance than only updating the structural features. This result is intuitive as a limitation of the original LBP is that the histogram can be sparse if some of the 256 LBP codes are not well represented. With the NLBP approach, the aggregation of the LBP encoding or bit map can be adjusted by updating the bin centers and widths to maximize the performance on different tasks such as image classification. For NLBP, all learning settings (\textit{i.e.}, updating structural and/or statistical textures) statistically significantly outperformed the baseline LBP ($71.66$ average test accuracy) feature and fixing all parameters demonstrating the utility of the proposed neural ``engineered" feature.

\subsection{Ablation Study: Neighborhood Size}
\label{sec:neighborhood}

{An important aspect of texture analysis is computing local features \mbox{\cite{peeples2021histogram}} and this is dependent on the neighborhood size. The results of this investigation are shown in Table \mbox{\ref{tab:hyperparams}} where we varied different dilation and aggregation window sizes for the statistical and structural textures, respectively. One observation is that the results for each method are relatively robust to the different combinations of hyperparameters with NEHD being more robust than NLBP. For NEHD, the size of the aggregation window for the histogram layer affected performance more than the increase in the dilation size of the edge kernel. NLBP was also more affected by the aggregation window size as in all cases of dilation, the $7 \times 7$ kernel had the lowest performance. These results show that the aggregation of statistical information can capture fine details of the structural features if the kernel size is not too large.}

\subsection{Ablation Study: Multichannel processing}
\label{sec:ablation_channels}
\begin{table}[htb]
    \centering
    \caption{Multichannel processing approaches for NEHD and NLBP on the PRMI dataset. The mean test accuracy with $\pm$ 1 standard deviation across five experimental runs is shown. The independent setting applied the neural ``engineered" feature to each channel separately, $1 \times 1$ convolution was a learnable mapping of the three channel input to a single channel, and grayscale was converting the image to a single channel intensity image. The NLBP and NEHD fusion approach with the best average test accuracy is bolded for each feature.}
    \begin{tabular}{|l|c|c|c|}
        \hline
        {Model} & {Independent} & $1 \times 1$ {Convolution} & {Grayscale} \\
        \hline
        NEHD & \textbf{89.92} $\pm$ \textbf{0.21} & $88.18 \pm 5.60$ & $89.45 \pm 1.56$ \\
        \hline
        NLBP & \textbf{91.17} $\pm$ \textbf{1.06} & $91.08 \pm 3.00$ & $88.76 \pm 6.44$ \\
        \hline
    \end{tabular}
    \label{tab:performance_configs}
\end{table}

\begin{table*}[htb]
    \centering 
    \caption{Comparison results of our proposed neural ``engineered" features, ``engineered" features, {and lightweight CNNs across} different datasets. The mean test accuracy with $\pm$ 1 standard deviation across multiple experimental runs is shown. The model with the highest test accuracy for each dataset is bolded.}
    \resizebox{0.5\textwidth}{!}{%
        \begin{tabular}{|l|c|c|c|}
            \hline
            \textbf{Method} & \textbf{FashionMNIST} & \textbf{PRMI} & \textbf{BloodMNIST } \\
            \hline
            EHD {\mbox{\cite{frigui2008detection}}}  & $86.94 \pm 0.00$ & $54.58 \pm 6.62$ & $66.33 \pm 0.07$ \\
            \hline
            LBP {\mbox{\cite{ojala1994performance}}}  & $71.66 \pm 0.02$ & $69.68 \pm 0.12$ & $43.25 \pm 0.07$ \\
            \hline
            LBP ROR {\mbox{\cite{ojala1996comparative}}} & $51.10 \pm 0.12$ & $70.70 \pm 0.11$ & $40.32 \pm 0.07$ \\
            \hline
            LBP Uniform {\mbox{\cite{ojala2002multiresolution}}} & $28.26 \pm 0.01$ & $67.17 \pm 0.00$ & $33.41 \pm 0.06$ \\
            \hline
            LBP NRI Uniform {\mbox{\cite{ahonen2004face,ahonen2006face}}} & $51.07 \pm 0.02$ & $71.30 \pm 0.03$ & $33.92 \pm 0.16$ \\
            \hline
            LBP Var {\mbox{\cite{ojala2002multiresolution}}} & $10.00 \pm 0.00$ & $59.93 \pm 0.11$ & $19.47 \pm 0.00$ \\
            \hline {DSANet\mbox{ \cite{bhowmik2024dsanet}}}  & $89.69 \pm 0.65 $ &  $45.78 \pm 7.59 $  & $86.43 \pm 2.04$ \\
            \hline
            {MSDCNN\mbox{ \cite{bhowmik2024utilization}}}  & $87.32 \pm 0.28$ &  $82.00 \pm 0.26 $  & \textbf{93.40} $\pm$ \textbf{0.05} \\
            \hline
            NEHD (ours) & \textbf{89.74} $\pm$ \textbf{0.12} & $89.92 \pm 0.21$ & 83.60 $\pm$ 0.35 \\
            \hline
            NLBP (ours) & $87.50 \pm 0.30$ & \textbf{91.17} $\pm$ \textbf{1.06} & $76.06 \pm 1.47$ \\
            \hline
        \end{tabular}%
    }
    \label{tab:method_comparison}
\end{table*}

\begin{figure*}[htb]
	\begin{subfigure}{.24\textwidth}{
			\includegraphics[draft=false,width=\textwidth]{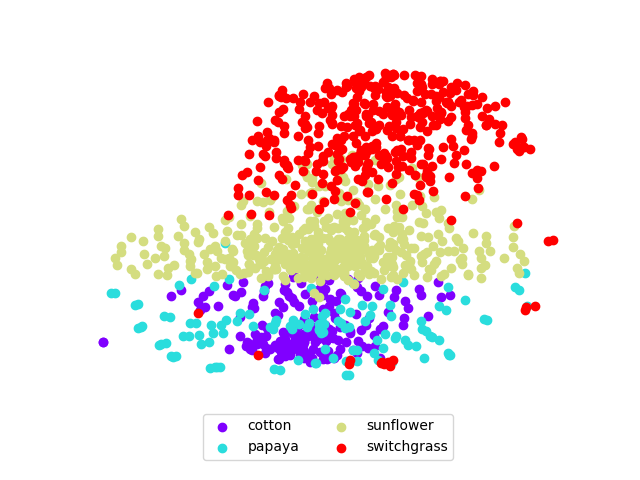}
			\caption{EHD (61.25\%)}
			\label{fig:TSNE_EHD}
		}
	\end{subfigure} 
	\begin{subfigure}{.24\textwidth}{
			\includegraphics[draft=false,width=\textwidth]{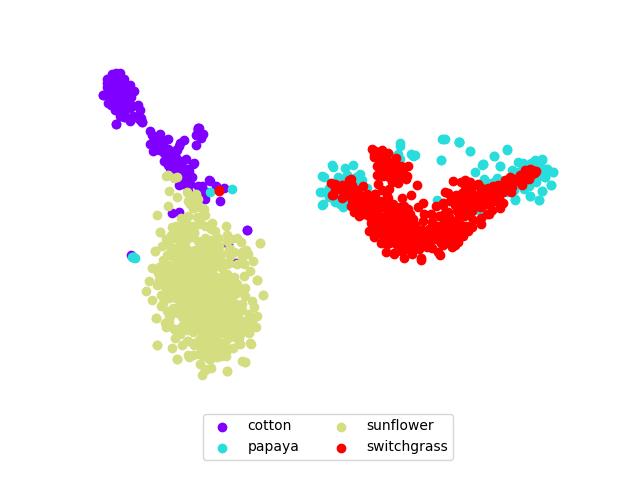}
			\caption{NEHD (90.17\%)}
			\label{fig:TSNE_NEHD}
		}
	\end{subfigure}
 	\begin{subfigure}{.24\textwidth}{
			\includegraphics[draft=false,width=\textwidth]{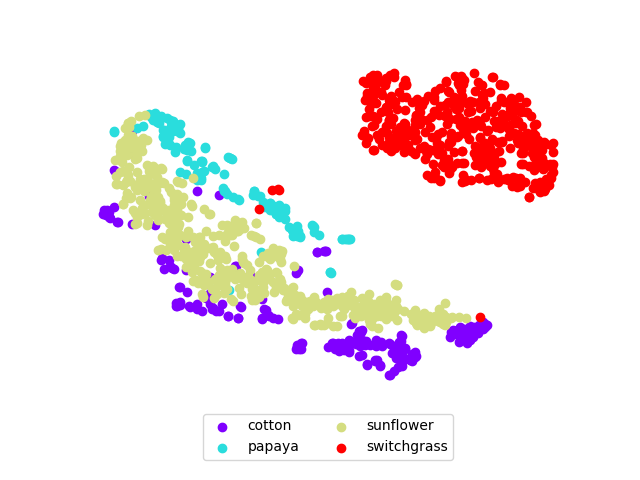}
			\caption{LBP (71.33\%)} 
			\label{fig:TSNE_LBP}
		}
	\end{subfigure} 
	\begin{subfigure}{.24\textwidth}{
			\includegraphics[draft=false,width=\textwidth]{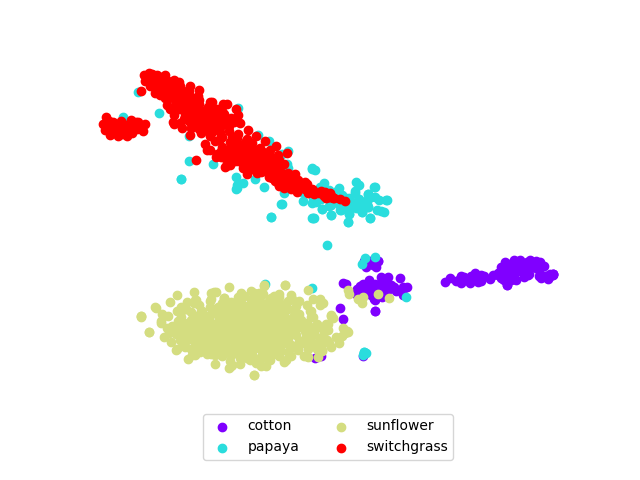}
			\caption{NLBP (92.50\%)}
			\label{fig:TSNE_NLBP}
		}
	\end{subfigure}
	\caption{t-SNE results for the best EHD, NEHD, LBP, and NLBP on the PRMI test set. The test accuracy for each approach is shown in parentheses. The random seed for each approach was the same in order to have a fair qualitative comparison between the feature visualizations. Each color corresponds to a different class in the PRMI dataset. The NEHD and NLBP models appear to have more compact classes in the projected space. For example, the switchgrass class (red data points) appears more compact for NEHD than EHD. When comparing NLBP and LBP, NLBP results in a more compact sunflower class (green data points).}
	\label{fig:TSNE}
\end{figure*} 

The results of our multichannel processing for the best NLBP and NEHD from Section \ref{sec:ablation_init} are shown in Table \ref{tab:performance_configs}. From the results, independently processing each channel achieved the highest average test accuracy. Independently processing each channel leads to the most information so this result is intuitive. However, the number of features will scale as the number of input feature channels are increased. When observing the results of converting to gray scale and applying a $1 \times 1$ convolution, the average test accuracy is comparable to the independent processing approach, although these approaches have increased variability indicated by the larger standard deviation when compared to the standard deviation of the independent processing approach. When integrating the neural ``engineered" features into a deeper network, the $1 \times 1$ approach may be the most ideal approach as this can reduce the number of features extracted when compared to the independent fusion approach. The grayscale conversion is limited to RGB inputs.

\subsection{Comparison of Neural and ``Engineered" Features}
\label{sec:comparison}
\begin{figure*}[htb]
	\begin{subfigure}{.245\textwidth}{
			\includegraphics[draft=false,width=\textwidth]{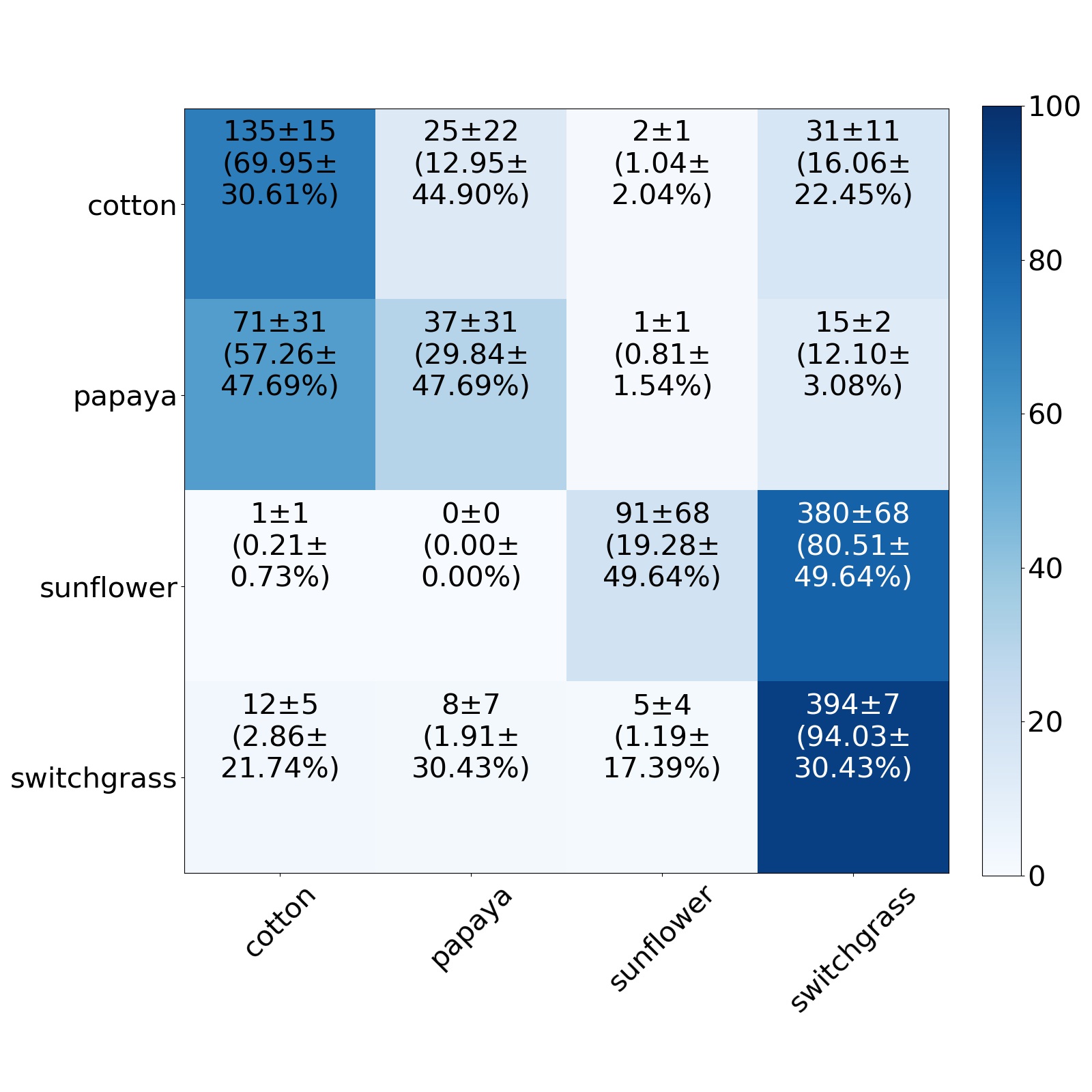}
            \caption{EHD ($54.58 \pm 6.62$\%)}
			\label{fig:Confusion_Matrix_EHD}
		}
	\end{subfigure} 
	\begin{subfigure}{.245\textwidth}{
			\includegraphics[draft=false,width=\textwidth]{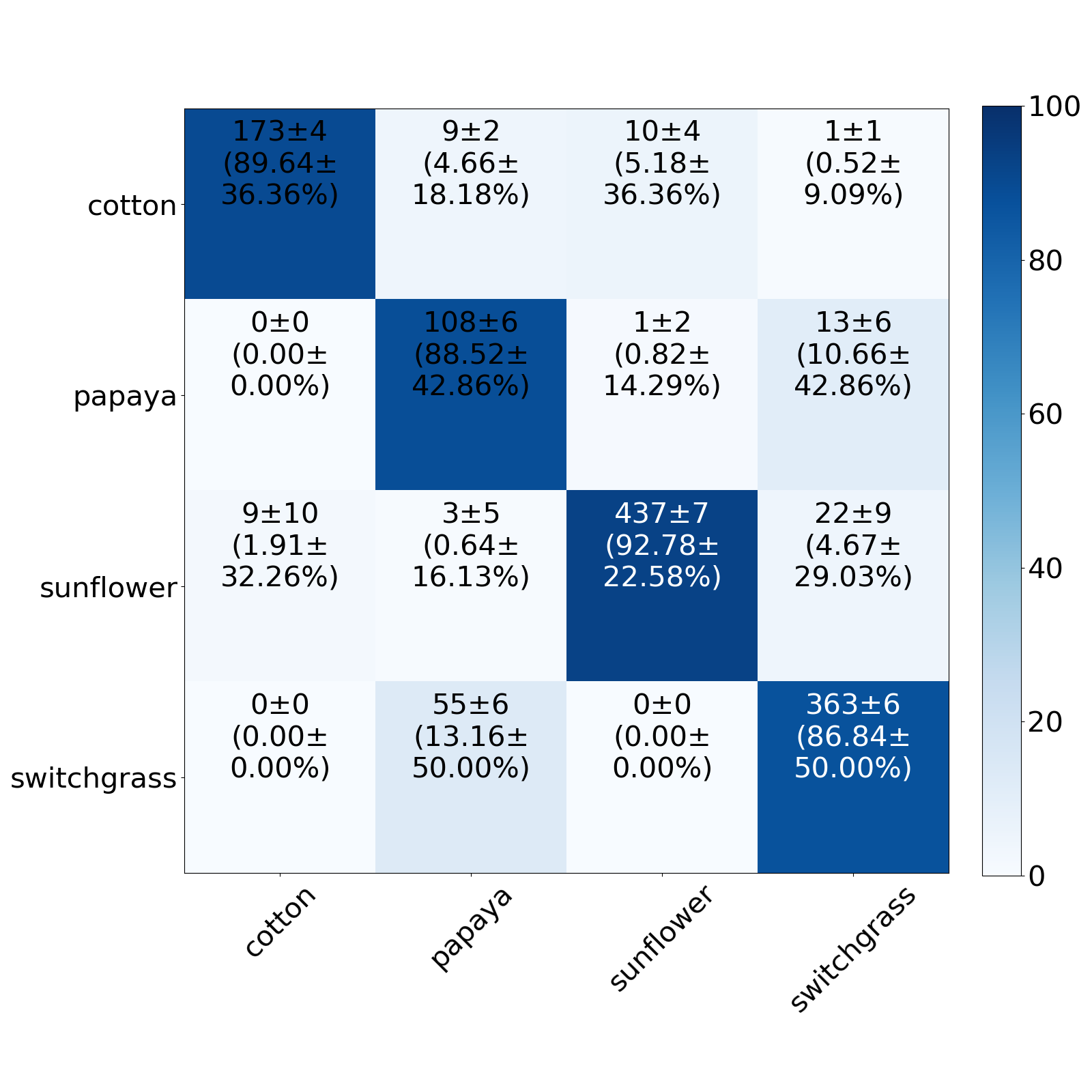}
            \caption{NEHD ($89.92 \pm 0.21$\%)}
			\label{fig:Confusion_Matrix_NEHD}
		}
	\end{subfigure}
 	\begin{subfigure}{.245\textwidth}{
			\includegraphics[draft=false,width=\textwidth]{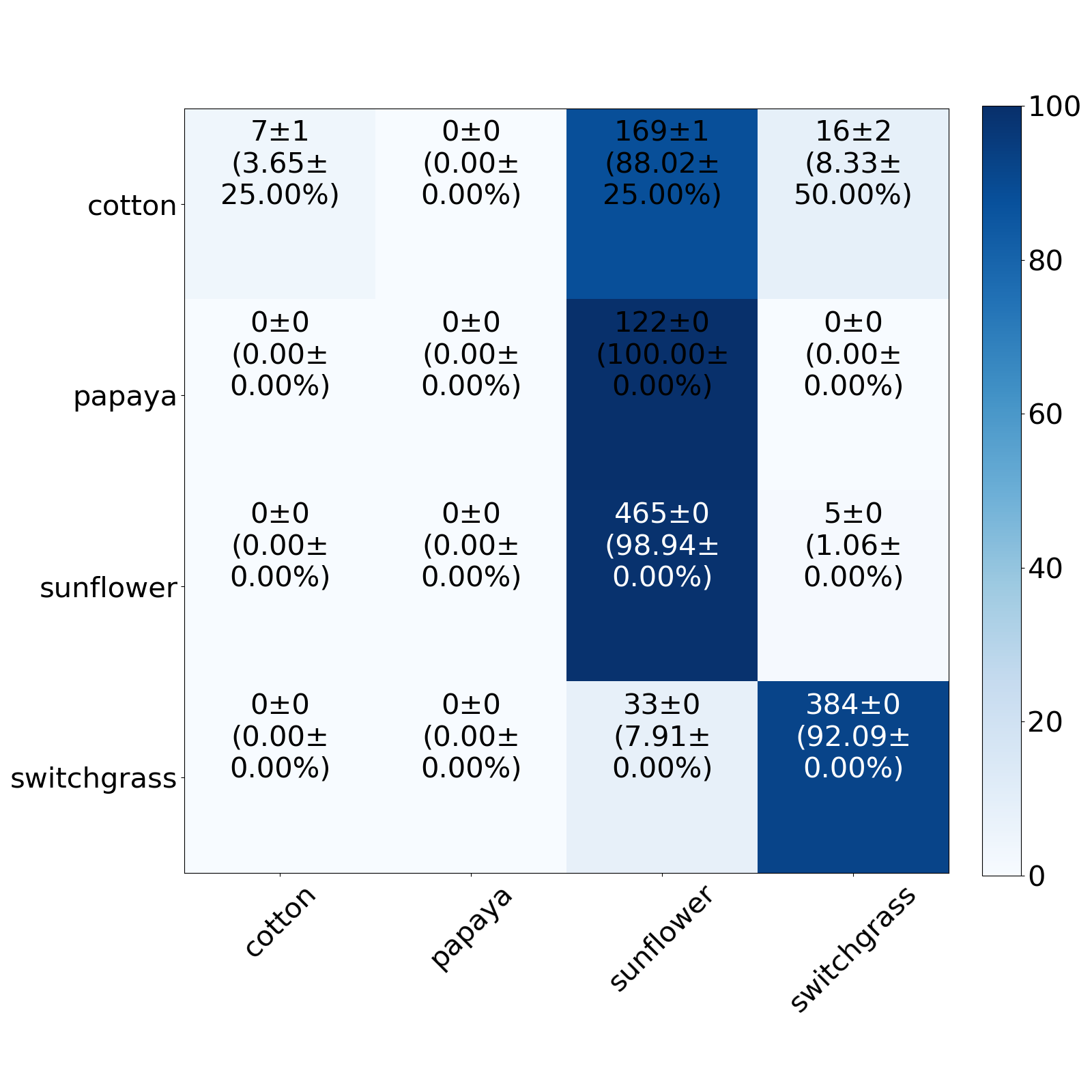}
            \caption{LBP ($71.30 \pm 0.03$\%)} 
			\label{fig: Confusion_Matrix_LBP}
		}
	\end{subfigure} 
	\begin{subfigure}{.245\textwidth}{
			\includegraphics[draft=false,width=\textwidth]{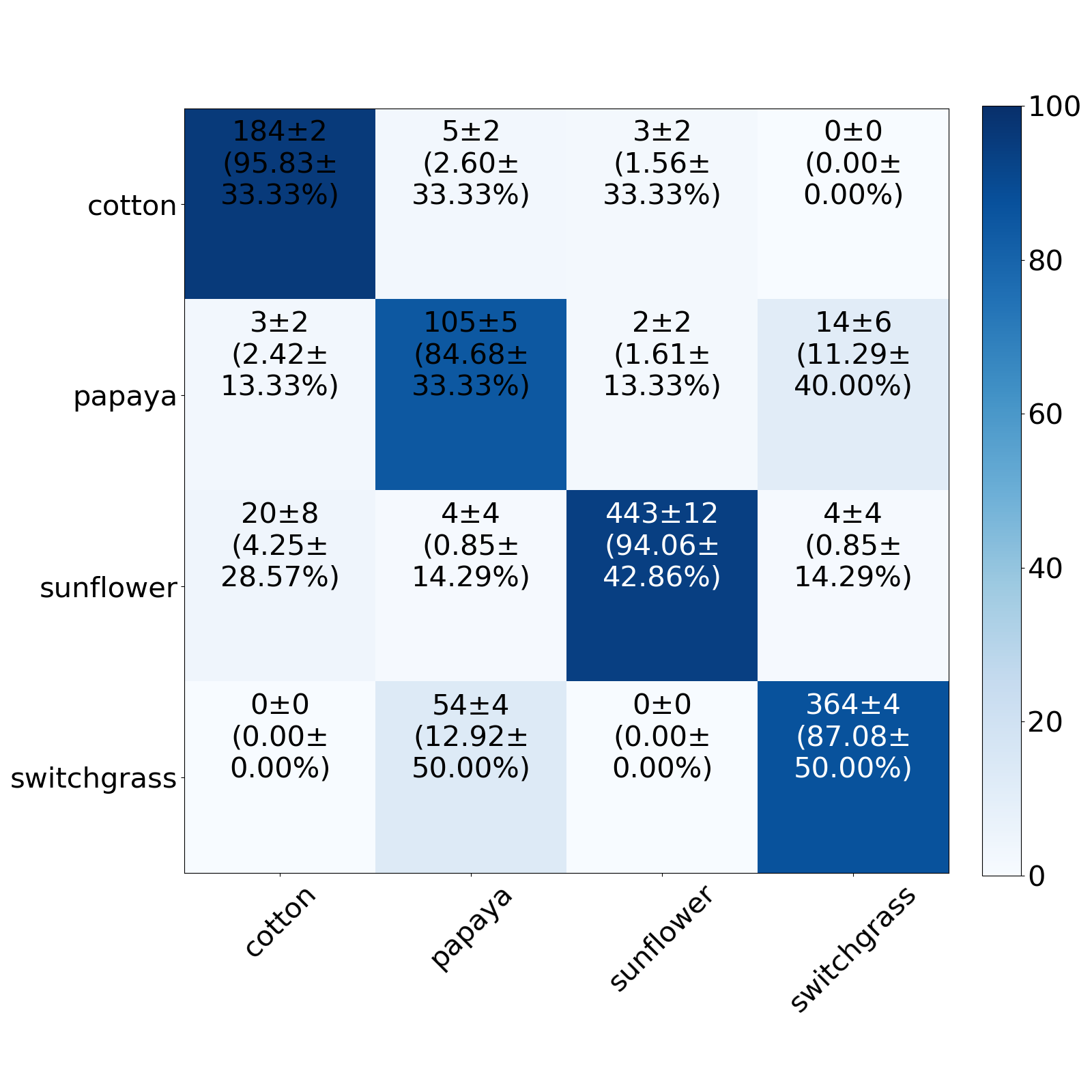}
            \caption{NLBP ($91.17 \pm 1.06$\%)}
			\label{fig:Confusion_Matrix_NLBP}
		}
	\end{subfigure}
	\caption{The confusion matrices for the best EHD, LBP (LBP NRI Uniform), NEHD, and NBLP are shown for the PRMI test set. The mean test accuracy with $\pm$ 1 standard deviation is shown in parentheses. Both NEHD and NLBP improve the identification of each class in comparison to the ``engineered" features.}
	\label{fig:conf_matrices}
\end{figure*} 

The last set of experiments compared our proposed NEHD and NLBP with the baseline EHD, LBP variants, and {lightweight CNNs} across three different datasets: FashionMNIST \cite{xiao2017fashion}, PRMI \cite{xu2022prmi}, BloodMNIST \cite{medmnistv1,medmnistv2,bloodmnist}. The results of the comparisons are shown in Table \ref{tab:method_comparison}. Overall, both NEHD and NLBP outperform the ``engineered" features across all three datasets. For FashionMNIST, as discussed in Section \ref{sec:ablation_init}, edges are important to distinguish the different articles of clothing resulting in EHD achieving the best performance for ``engineered" features. However, our NEHD and NLBP achieve statistically significantly (\textit{i.e.}, no overlapping error bars) improved performance. {When compared with the DSANet \mbox{\cite{bhowmik2024dsanet}} and MSDCNN \mbox{\cite{bhowmik2024utilization}}, our proposed NEHD outperforms these two models on this datasets and NLBP performs better than MSDCNN as well. This shows the power of these learnable histogram-based texture features can potentially capture useful information more directly than deeper CNNs. Also, our proposed models have significantly less learnable parameters than the lightweight CNNs. For FashionMNIST, DSANet and MSDCNN had a total of 290,894 and 630,104 respectively. The proposed NEHD and NLBP have 43,732 and 92,282 respectively which further supports the potential of these neural ``engineered" features. If only the feature extraction layers learnable parameters are considered, NEHD only has 162 while NLBP has 112.}

PRMI and BloodMNIST are consists of RGB images. Each feature was applied to each channel independently as a fair comparison between each approach. The PRMI results show that NLBP slightly outperformed NEHD. For the PRMI dataset, edge information is important to identify the roots, but there are some images have some illumination differences. LBP is robust to illumination variations (\textit{e.g.}, monotonic changes) \cite{humeau2019texture}, resulting in NLBP retaining a similar property when extracting texture features from the images. {NEHD and NLBP also performed well on the PRMI dataset in comparison to the DSANet and MSDCNN. The difference in performance of the MSDCNN vs DSANet may be due to the multi-scale convolutional approach employed in the network. However, the results on the PRMI further support the usefulness of texture features especially in Agriculture focused applications \mbox{\cite{mohan2024lacunarity}}.}

The NEHD and NLBP are compared with the best EHD and LBP variant qualitatively (Figure \ref{fig:TSNE}) and quantitatively (Figure \ref{fig:conf_matrices}). As seen in the TSNE visuals in Figure \ref{fig:TSNE}, the features learned by NEHD and NLBP appear more compact in the 2D projection than the ``engineered" features. An important aspect of both approaches is the aggregation performed by the histogram layer. The soft binning approach provides a way to account for intra-class variations that can be present within the dataset \cite{peeples2021histogram}. As noted in the confusion matrices in Figure \ref{fig:conf_matrices}, the neural ``engineered" features outperform the ``engineered" features. For some of the PRMI images, the roots are occluded with background information (\textit{e.g.}, soil, image artifacts) and the fixed representation of the ``engineered" features is limited in mitigating the impact of these confusers in the images. However, our neural ``engineered" features can change the texture representation to improve classification performance. 

BloodMNIST was the most difficult dataset for the shallow networks used since the number of classes was eight (more than PRMI) in comparison and the blood cells visually have small differences between the different classes. However, our proposed features achieved statistically significant higher test accuracy than the ``engineered" features. Texture information is vital in several biomedical applications \cite{liu2019bow}. Our proposed approach can be applied to several biomedical tasks and integrated into deeper networks to achieve improved performance on the BloodMNIST dataset. {The lightweight CNNs did perform better on this dataset than NEHD and NLBP. As discussed, this was the most challenging dataset of the three and deeper networks will improve the feature representations learned for classifying the blood cells. An interesting observation is that the multi-scale network (MSDCNN) performed the best on this datasets. Future studies could investigate adapting NEHD and NLBP for multi-scale texture analysis especially since texture is very scale dependent \mbox{\cite{mohan2024lacunarity,petrou2021image}}.}

\section{Conclusion} In this work, we proposed neural ``engineered" features by using histogram layers to aggregate structural texture features. We demonstrate the approach by introducing two neural ``engineered" features: NEHD and NLBP. Our results across benchmark (FashionMNIST) and real-world (PRMI and BloodMNIST) datasets showcase the potential use of these features. The general framework for neural ``engineered" features can be used for other texture feature approaches such as Haralick texture features \cite{haralick1973textural}, histogram of oriented gradients \cite{dalal2005histograms} and several more statistical and structural texture feature extraction methods \cite{liu2019bow,humeau2019texture}. Future work includes integrating the neural handcrafted layer(s) into deeper networks, improving the multichannel processing approach, and designing new objective functions to maximize statistical and/or structural texture information.


\section*{Acknowledgment}
This material is based upon work supported by the National Science Foundation Graduate Research Fellowship under Grant No. DGE-1842473 and by the Office of Naval Research grant N00014-16-1-2323. The views and opinions of authors expressed herein do not necessarily state or reflect those of the United States Government or any agency thereof. Portions of this research were conducted with the advanced computing resources provided by Texas A\&M High Performance Research Computing.


\printbibliography

\begin{IEEEbiography}[{\includegraphics[width=1in,height=1.25in,clip]{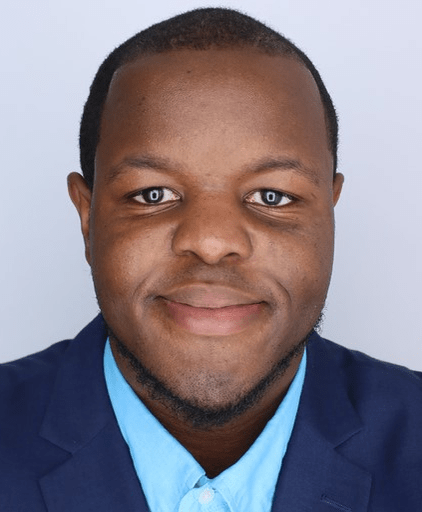}}]{Joshua Peeples} (Member, IEEE)
received the Ph.D. degree in electrical and computer engineering from the University of Florida in 2022. He is an Assistant Professor in the Department of Electrical and Computer Engineering at Texas A\&M University. His primary research interests include machine learning, computer vision, and image processing with a focus on image texture analysis.
\end{IEEEbiography}

\begin{IEEEbiography}[\vspace{-7mm}{\includegraphics[width=1in,height=1.25in,clip,keepaspectratio]{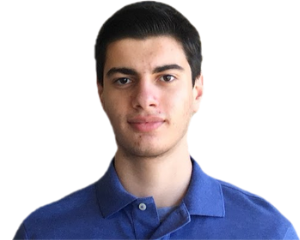}}]{Salim Al Kharsa}
is a part-time Computer Engineering graduate student at Texas A\&M University, where his research focuses on developing lightweight and reproducible evaluation toolkits for generative AI systems, particularly large language and vision-language models. He works full-time as a machine learning engineer with a focus on real-time modeling and predictive analytics.
\end{IEEEbiography}

\begin{IEEEbiography}[\vspace{-5mm}{\includegraphics[width=1in,height=1.25in,clip,keepaspectratio]{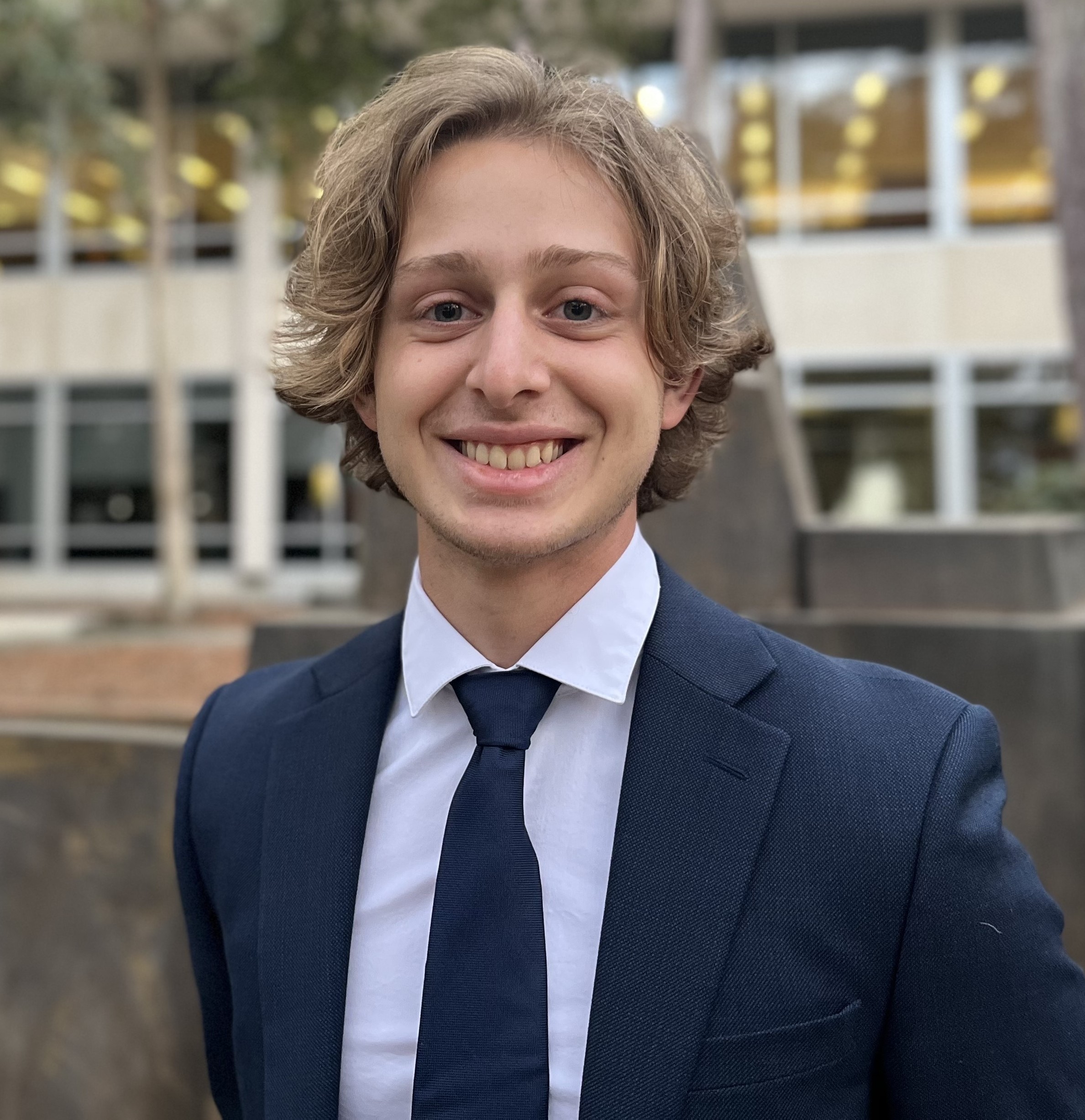}}]{Luke Saleh} (Student Member, IEEE)
is a graduate Computer Engineering student at the University of Florida (UF) with an expected graduation date of December 2025 with a Master of Science in Electrical and Computer Engineering. He is currently working as a graduate researcher for the Machine Learning and Sensing Lab at UF. His primary interests include embedded systems, machine learning, digital hardware design, and data science. 
\end{IEEEbiography}

\begin{IEEEbiography}
[\vspace{-5mm}{\includegraphics[width=1in,height=1.25in,clip,keepaspectratio]{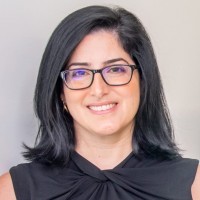}}]{Alina Zare} (Senior Member, IEEE) teaches and conducts research in the area of machine learning and artificial intelligence as a Professor in the Electrical and Computer Engineering Department at the University of Florida. She also serves as the Director for the campus-wide Artificial Intelligence and Informatics Research Institute. Dr. Zare’s research has focused primarily on developing new machine learning algorithms to automatically understand and process data and imagery. Her research work has included automated plant root phenotyping, sub-pixel hyperspectral image analysis, target detection and underwater scene understanding using synthetic aperture sonar, LIDAR data analysis, Ground Penetrating Radar analysis, and buried landmine and explosive hazard detection.
\end{IEEEbiography}

\end{document}